\documentclass[10pt,twocolumn,letterpaper]{article}

\usepackage[final]{cvpr}      

\usepackage{graphicx}
\usepackage{amsmath}
\usepackage{amssymb}
\usepackage{booktabs}
\usepackage{multirow}

\usepackage[pagebackref,breaklinks,colorlinks]{hyperref}

\usepackage[capitalize]{cleveref}
\crefname{section}{Sec.}{Secs.}
\Crefname{section}{Section}{Sections}
\Crefname{table}{Table}{Tables}
\crefname{table}{Tab.}{Tabs.}

\begin{document}

\title{CPA: Camera-pose-awareness Diffusion Transformer for Video Generation}

\author{YueLei Wang\textsuperscript{\rm 1}~~~ 
    Jian Zhang\textsuperscript{\rm 1} ~~~
    Pengtao Jiang\textsuperscript{\rm 1}~~~ 
    Hao Zhang\textsuperscript{\rm 1}~~~
    Jinwei Chen\textsuperscript{\rm 1}~~~
    Bo Li\textsuperscript{\rm 1} \\
\textsuperscript{\rm 1}Image Algorithm Research Department, vivo Mobile Communication Co., Ltd\\ 
}

\maketitle

\begin{abstract}
   Despite the significant advancements made by Diffusion Transformer (DiT)-based methods in video generation, there remains a notable gap with controllable camera pose perspectives. Existing works such as OpenSora do NOT adhere precisely to anticipated trajectories and physical interactions, thereby limiting the flexibility in downstream applications. To alleviate this issue, we introduce CPA, a unified camera-pose-awareness text-to-video generation approach that elaborates the camera movement and integrates the textual, visual, and spatial conditions. Specifically, we deploy the Sparse Motion Encoding (SME) Module to transform camera pose information into a spatial-temporal embedding and activate the Temporal Attention Injection (TAI) Module to inject motion patches into each ST-DiT block. Our plug-in architecture accommodates the original DiT parameters, facilitating diverse types of camera poses and flexible object movement. Extensive qualitative and quantitative experiments demonstrate that our method outperforms LDM-based methods for long video generation while achieving optimal performance in trajectory consistency and object consistency.
\end{abstract}

\section{Introduction}
\label{sec:intro}

\begin{figure}
  \centering
  
   \includegraphics[width=0.9\linewidth]{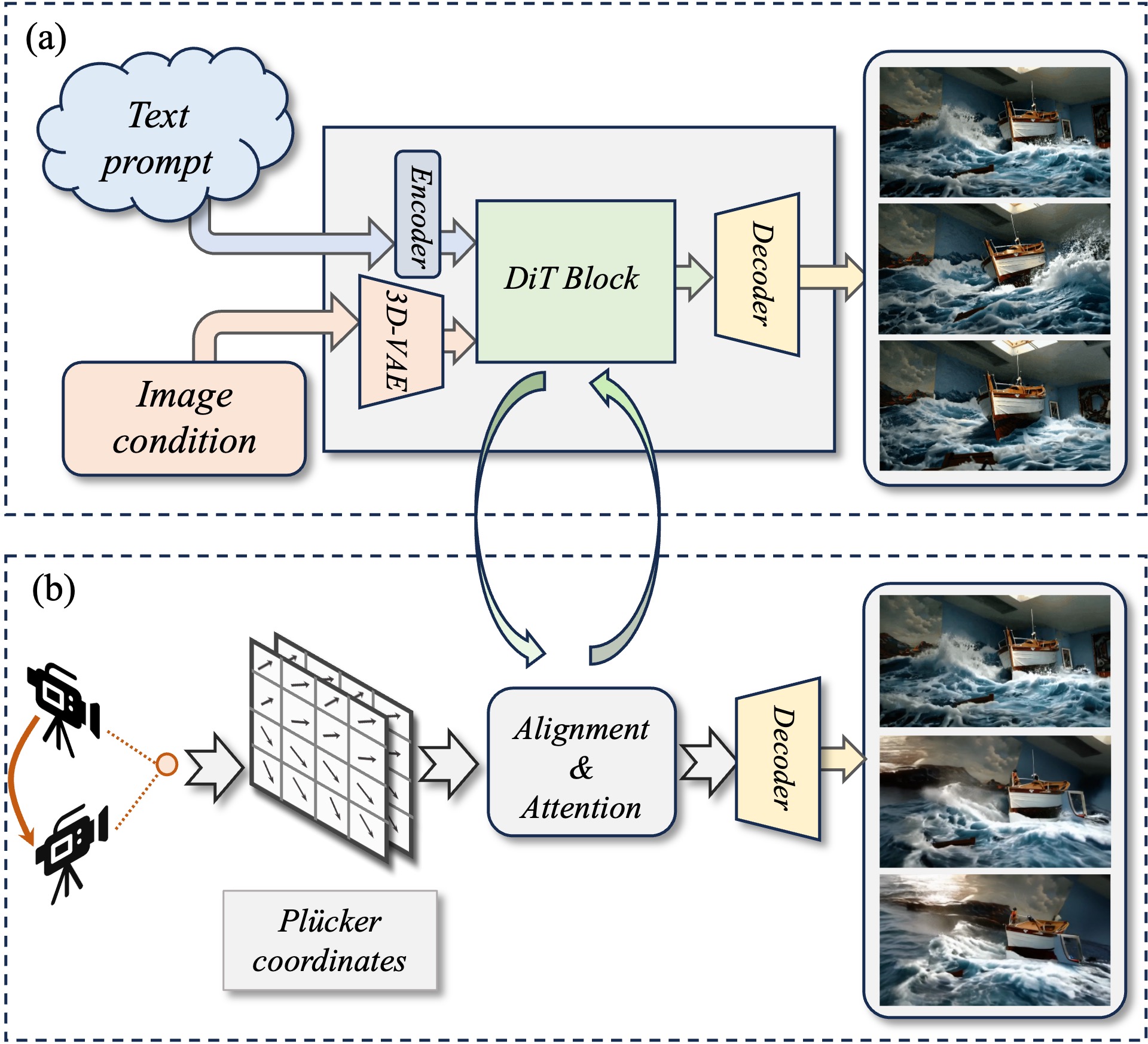}

   \caption{The relevance of this work to video generation models. (a) The DiT-based video generation model leverages DiT blocks to produce high-quality videos. (b) CPA utilizes Plücker coordinates encoded with camera pose information and aligns with the attention mechanism in the DiT block to generate camera-oriented videos.
}
   \label{fig:intro}
\end{figure}

The rapid evolution of video generation has been characterized by the rise of the DiT method~\cite{peebles2023scalable}, which is indispensable for effective long-sequence training and low-latency inference.  Despite these advancements, DiT models struggle with controllability, especially concerning the precise modulation of camera movements — a critical technique for numerous creative applications.

Recent prominent text-to-video (T2V) approaches such as AnimateDiff~\cite{guo2023animatediff}, Lumiere~\cite{bar2024lumiere}, and SVD~\cite{blattmann2023stable}, incorporate personalized text-to-image (T2I) models and further modify the U-Net architecture ~\cite{ronneberger2015u} by introducing temporal embeddings and spatial-temporal cross-attention to ensure consistency across frames. Currently, taking into account the global property of camera motion and the local property of object motion information, MotionCtrl~\cite{wang2024motionctrl} and CameraCtrl~\cite{he2024cameractrl} significantly enhance the possibilities of fine-grained content generation. However, these methods are practically constrained by the Latent Diffusion Models (LDM)~\cite{rombach2022high}, which imposes strict limitations on the latent space. Evidence shows that the U-Net architecture struggles to accommodate variations in video resolution and duration due to preset constraints on temporal length and dimensions of latent space, which limit its ability to extend frame number or higher resolution. With the release of Sora~\cite{videoworldsimulators2024} earlier this year, DiT-based frameworks demonstrate remarkable proficiency in producing high-quality and long-term video content. On the one hand, recent works such as Kling, OpenSora~\cite{opensora}, and Open-Sora-Plan~\cite{pku_yuan_lab_and_tuzhan_ai_etc_2024_10948109}  conduct extensive explorations on 3D-VAE and spatial-temporal DiT (ST-DiT), achieving promising results in the T2V task. On the other hand, for applications concentrating more on motion manipulation, Tora~\cite{zhang2024tora} implements the extraction of object trajectory data into motion-guided fusion, thereby enabling scalable and flexible video generation. However, an effective solution for enhancing controllable video generation with camera pose sequences remains elusive, even ignored. Compared to object trajectories, camera pose requires more complex motion matrices, making it challenging to incorporate this task into a Transformer framework with variable frame numbers.

Therefore, we propose a \textbf{c}amera-\textbf{p}ose-\textbf{a}wareness approach for DiT-based video generation (\textbf{CPA}), which addresses the problem of precise control over camera pose sequences while preserves the intrinsic visual quality and extrinsic object movement, as depicted in Fig.~\ref{fig:intro}. Our method utilizes the OpenSora-v1.2 framework and extracts inter-frame motion sequences from reference videos in camera perspectives. First, each frame is annotated with a 12-dimension motion matrix, including a $3\times3$ rotation matrix and a $3\times1$ translation matrix. Effectively capturing the precision of the camera pose remains a challenge. We propose the Sparse Motion Encoding Module for converting camera rotation and translation parameters into a sparse motion field based on Plücker coordinates. Second, The Temporal Attention Injection Module is used to align the camera pose latent with the temporal attention features, through layer normalization and MLP. Furthermore, a VAE~\cite{kingma2013auto} is trained for the reconstruction of camera pose latent space, improving its alignment with the temporal attention layer.

The training of CPA consists of two parts. First, the reconstruction loss is adopted for the camera pose sequences during VAE training. We pick RealEstate10K~\cite{zhou2018stereo}, a video dataset with over 60k camera pose annotations, to train the VAE for encoding the aforementioned sparse motion field. Second, we fine-tune the OpenSora by freezing all layers except temporal attention layers, retaining the initial capabilities of the model while effectively injecting camera information.  We evaluate our method and the experiments show that our approach achieved state-of-the-art~(SOTA) performance for long video generation tasks.

\begin{figure*}
  \centering
  
   \includegraphics[width=0.8\linewidth]{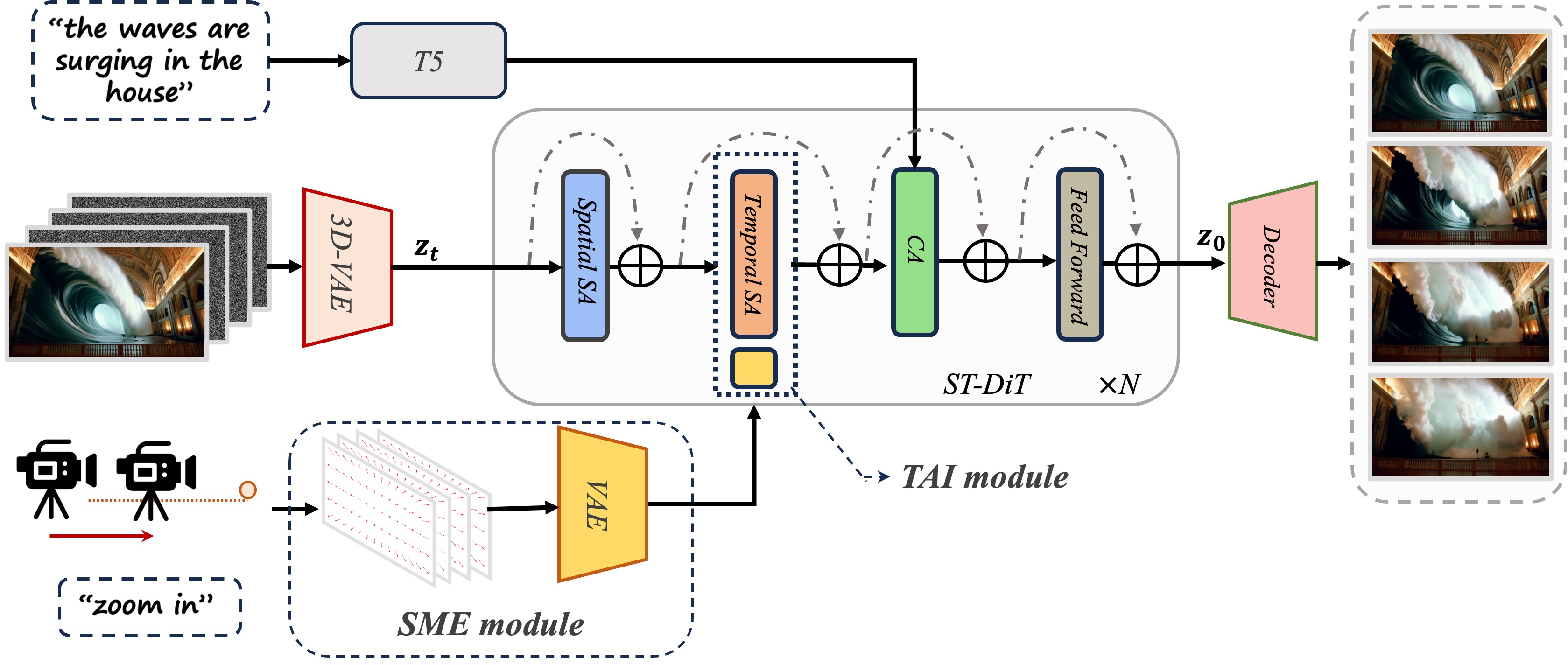}

   \caption{The overview of CPA. CPA includes the Sparse Motion Encoding~(SME) Module and the Temporal Attention Injection~(TAI) Module. It establishes a sparse motion sequence representation based on Plücker coordinates and feeds it into the VAE for pose latent, handling the camera pose sequences for multiple frames. By employing layer normalization and MLP, it achieves alignment of the temporal attention layer and the pose latent. The inputs of the video and text caption are consistent with OpenSora, feeding into the ST-DiT and cross-attention layers through the 3D-VAE and T5 models, respectively.}
   \label{fig:overview}
\end{figure*}

Our main contributions are:
\begin{itemize}
    \item We introduce CPA, empowering diffusion transformer with precise control over camera pose. A mathematical derivation is consolidated for embedding the camera intrinsic and extrinsic parameters to the motion field based on Plücker coordinates, easing the burden of capturing minor perturbations of camera pose.
    \item  We propose two plug-in modules: Sparse Motion Encoding Module and Temporal Attention Injection Module, which compacts the extracted camera pose embedding and effectively integrates it with the framework.
    \item  Extensive experiments and comprehensive visualization demonstrate that our method achieves a promising camera-instruction following capability while maintaining the high-fidelity object appearance. 
\end{itemize}

\section{Related Work}

\subsection{Video Generation}
With diffusion models being proven as an effective method for creating high-quality images, research on dynamic video generation has gradually emerged. Make-a-video~\cite{singer2022make} and MagicVideo~\cite{zhou2022magicvideo} use 3D U-Net in LDM to learn temporal and spatial attention, though the training cost is relatively expensive. VideoComposer~\cite{wang2024videocomposer} expands the conditional input forms by training a unified encoder. Other methods (Align Your Latents~\cite{blattmann2023align}, VideoElevator~\cite{zhang2024videoelevator}, AnimateDiff, Direct a Video~\cite{yang2024direct}, Motioni2v~\cite{shi2024motion}, Consisti2v~\cite{ren2024consisti2v}) improve the performance by reusing T2I models and make adjustments in the temporal and spatial attention parts to reduce issues such as flicker reduction. Video generation models based on DiT or Transformer~\cite{vaswani2017attention} adopt spatial-temporal attention from LDM, such as Latte~\cite{ma2024latte}, Vidu~\cite{bao2024vidu}, CogVideoX~\cite{yang2024cogvideox} and SnapVideo~\cite{menapace2024snap}, which have significant advantages in terms of resolution and duration compared to LDM methods.

\subsection{Controllable Generation}

Controllable generation is one of the key research topics for generative tasks. For T2I task, ControlNet~\cite{zhang2023adding} enables fine-tuning samples while retaining the backbone, and ControlNeXT~\cite{peng2024controlnext} significantly improves training efficiency. For controllable video generation, tune-a-video~\cite{wu2023tune} enables single sample fine-tuning, changing styles while maintaining consistent object motion. MotionClone~\cite{ling2024motionclone} implements a plug-and-play motion-guided model. MotionCtrl and CameraCtrl use motion consistency modules to introduce camera pose sequences. PixelDance~\cite{zeng2024make} uses the first and the last frame as a reference for video generation. Image Conductor~\cite{li2024image} and FreeTraj~\cite{qiu2024freetraj} introduce tracking schemes based on trajectories and bounding boxes, respectively. ViewDiff~\cite{hollein2024viewdiff} reconstructs 3D information of objects based on camera pose sequences. 
Nevertheless, the aforementioned methods struggle with sustaining continuous and consistent control within long-form videos, a challenging issue owing to the intrinsic capacity and scalability limitations of the U-Net design. In parallel, diffusion Transformer demonstrate the feasibility of generating high-fidelity long videos while scarce research above DiT is concentrated on precise camera pose control.
VD3D builds on SnapVideo, embedding camera pose into cross-attention layers via Plücker coordinates. Tora and TrackGo~\cite{zhou2024trackgo} explore controllable video generation by trajectories and masks. Currently, there is still limited work for camera pose information on DiT.

\section{Method}
\subsection{Preliminary}
The LVDM (Latent Video Diffusion Model)~\cite{he2022latent} aims to video generation through a denoising diffusion network like U-Net. It proposes a strategy for the separation of spatiotemporal self-attention to address the frame motion coherence in video generation. The loss function for the U-Net is shown in the following formula:
\begin{equation}
   \mathcal{L}(\theta) = \mathbb{E}_{z_0,c,t,\epsilon  }[\left\| \epsilon_\theta (z_t,c,t) - \epsilon \right\|_2 ^2]
  \label{eq:loss}
\end{equation}
Here,  $\epsilon_\theta$ is the predicted noise,  $z_t$ and  $c$ represent the latent space at $t$ step and text condition, respectively. The latent space of the U-Net conforms to the following Markov chain:
\begin{equation}
   z_t = \sqrt{\bar{\alpha_t}}z_0 + \sqrt{1-\bar{\alpha_t}} \epsilon 
  \label{eq:zt}
\end{equation}
where $\bar{\alpha_t}=\prod _{i=1}^t \alpha_t$, $\alpha_t$ represents the noise strength in step $t$.

The DiT-based method replaces the U-Net with Transformer, remaining its sequential processing capabilities to greatly enhance the image quality and duration in video generation. To reduce computational complexity, the 3D-VAE in OpenSora performs a 4$\times$ compression on the temporal dimension. Compared to LVDM's latent space of $b \times L \times w \times h$, OpenSora's latent space size is $ b\times f\times w\times h(f=L/4)$, which is more lightweight on the temporal dimension.

\subsection{CPA}
\label{sec:CPA}

\begin{figure*}
  \centering
  
   \includegraphics[width=0.8\linewidth]{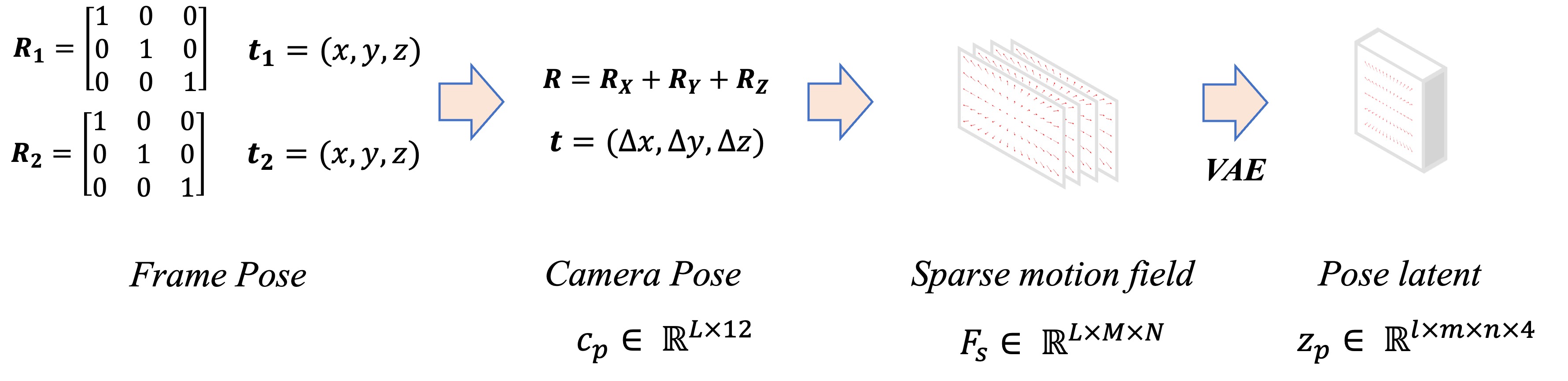}

   \caption{The pipeline of camera pose sequences encoding. The matrix parameters between adjacent frames are calculated to obtain the camera pose sequence, which is then transformed into RGB space through the sparse motion field and finally processed into pose latent by the VAE.
}
   \label{fig:pose}
\end{figure*}

As depicted in Fig.~\ref{fig:overview},  the proposed CPA consists of two modules: the Sparse Motion Encoding Module and the Temporal Attention Injection Module. First, an explanation of the calculation of Plücker coordinates is provided, which can provide more detailed information than directly using the camera pose. Subsequently, the detailed optimizations for the module will be presented.

The representation of a 2D image $\mathbf{x}$ requires a projection transformation $\mathbf{P}$ based on real-world 3D coordinates. For a point $\mathbf{X}=[X,Y,Z,1]^T$  in the 3D world, this transformation is typically achieved using a rotation matrix $\mathbf{R}$ combined with a translation component $\mathbf{t}$ represented as follows.

\begin{equation}
    \mathbf{x} = \mathbf{P} \mathbf{X} = \begin{bmatrix} \mathbf{R} \mid  \mathbf{t} \end{bmatrix} \mathbf{X}
    \label{eq:3d2d}
\end{equation}

Therefore, for the camera coordinate, we have $\mathbf{x}_c = \mathbf{R} \mathbf{X} + \mathbf{t}$. By introducing the camera intrinsic matrix $\mathbf{K}$, the 2D image point can be mapped to pixel coordinates:

\begin{equation}
    \mathbf{x} = \mathbf{K} \mathbf{x}_c = \mathbf{K} (\mathbf{R} \mathbf{X} + \mathbf{t})
    \label{eq:cam}
\end{equation}

To back-project the 2D image coordinates to camera coordinates, we use the  camera intrinsic matrix $\mathbf{X}_{\text{img}} = \mathbf{K}^{-1} \begin{bmatrix} x,y,1 \end{bmatrix}^T$. Then, using approach similar to the transition from Equation~\ref{eq:3d2d} to Equation~\ref{eq:cam}, the coordinate transformation formula for camera coordinate $\textbf{Q}_{x,y}$ is:

\begin{equation}
    \textbf{Q}_{x,y}= \textbf{RK}^{-1}\left [ x,y,1 \right ]^T + \textbf{t} 
  \label{eq:3dpoint}
\end{equation}

By introducing the homogeneous coordinates $\begin{bmatrix} \mathbf{o}_{c}, 1 \end{bmatrix}$, where $\mathbf{o}_{c}$ represents the optical camera center. The final equation is following:

\begin{equation}
    \textbf{P}_{x,y}=\left [ \textbf{o}_c,1 \right ]\left ( \textbf{RK}^{-1}\left [ x,y,1 \right ]^T + \textbf{t} \right )
  \label{eq:plucker}
\end{equation}

We can get  Plücker coordinates~\cite{pvribyl2016camera}, which is used in ~\cite{he2024cameractrl,bahmani2024vd3d}. By calculating between adjacent frames, it forms the motion vector from the camera center to the camera coordinate $(x,y)$.
For the method of directly using motion matrices ~\cite{wang2024motionctrl}, camera poses are serialized frame-by-frame into $ c_p\in \mathbb{R}^{L\times 12}$, where $L$ denotes the frame number. During motion injection,  the parameters are replicated in spatial dimensions to align temporal attention layer. However, this approach may encounter problems with the DiT-based method that exists in time-dimensional compression.

\textbf{Sparse Motion Encoding Module.} In this work, we propose a method for converting a pixel-wise motion field based on Plücker coordinates into a sparse motion field, as shown in Fig.~\ref{fig:pose}. Although Plücker coordinates can precisely describe the motion trajectory for each pixel in the image, we perform sparse sampling of the motion field to enhance computational efficiency and adapt to spatial domain feature representation. Assuming the image resolution is  $W \times H$, we sample every  $s_x$  pixels in the  $x$  direction and every  $s_y$  pixels in the $y$ direction to obtain a sparse point sequence $\{(x_i, y_j)\}$, with the corresponding sparse motion trajectory given by:

\begin{equation}
\mathbf{P}_{x_i,y_j} = \left[ \mathbf{o}_c, 1 \right] \left ( \mathbf{R}\mathbf{K}^{-1} \left [ x_i,y_j,1 \right ]^T + \mathbf{t}
  \label{eq:plucker1} \right )
\end{equation}
where  $x_i = i \cdot s_x$  and  $y_j = j \cdot s_y$ , with  $i$  and $j$   being the sampling indices. Here, we get a sparse motion field $F_s \in \mathbb{R}^{L\times M \times N}$, the $M=W / s_x$, $N=H / s_y$.

We train a VAE to compress the sparse motion field, aligning it with the temporal sequences in OpenSora. MagViT-v2~\cite{yu2023language} is selected to maintain consistency with the temporal attention layers and the reconstruction loss of the camera pose motion is calculated. We get the pose latent $z_p \in \mathbb{R}^{l\times m \times n \times 4}$, where $l = L / 4$,$m = M / 8$, $n = N / 8$.

\begin{figure}
  \centering
  
   \includegraphics[width=0.6\linewidth]{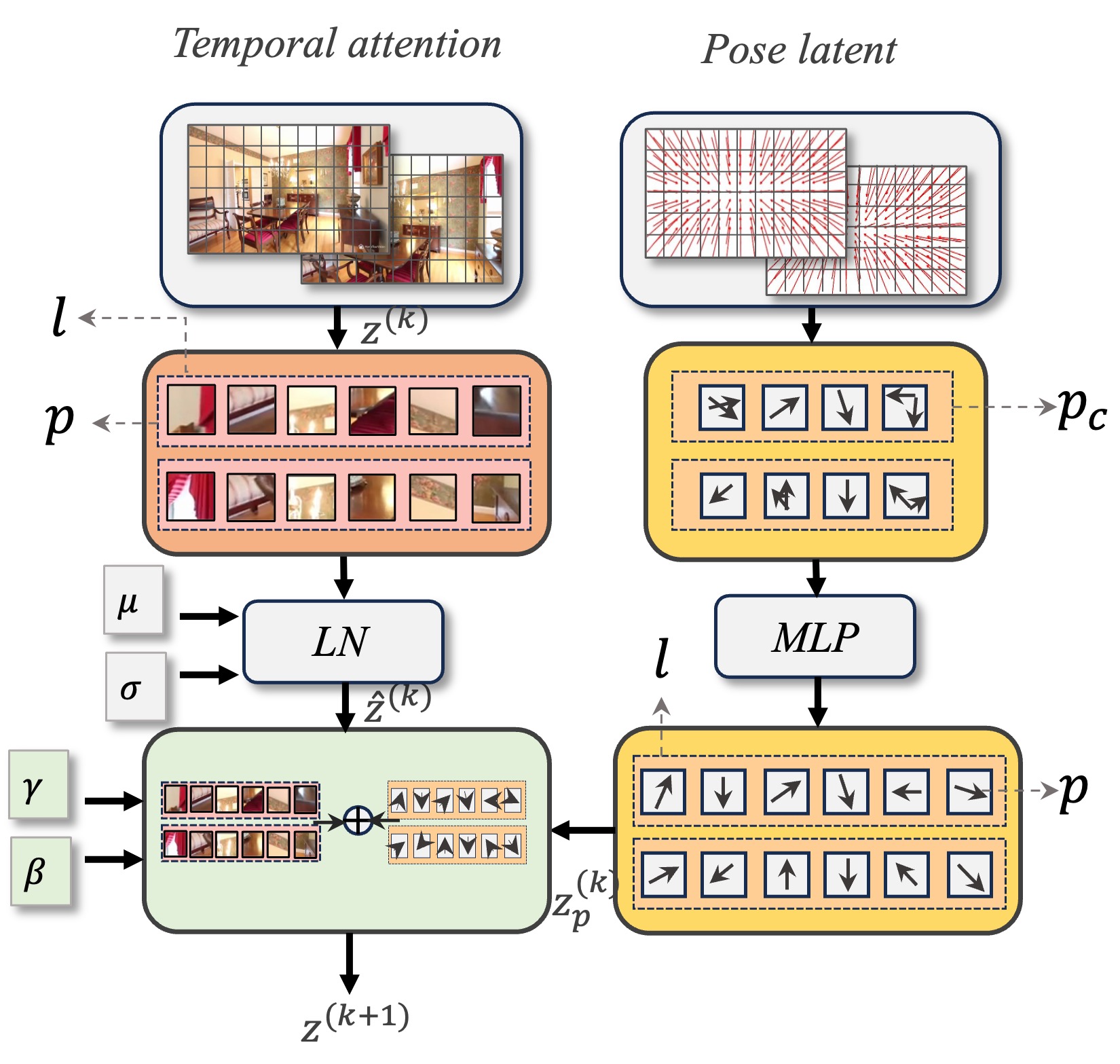}

   \caption{Temporal Attention Injection Module. Layer normalization~(LN) and multi-layer perceptron~(MLP) are used during processing temporal attention features and pose latent orientation, respectively.}
   \label{fig:TAI}
\end{figure}

\textbf{Temporal Attention Injection Module.} As shown in Fig.~\ref{fig:TAI}, we use layer normalization and MLP to align pose latent with temporal attention layer. The pose latent after the SME  has $l$ layers, and each layer has $p_c=m \times n$ patches, while for temporal attention layers, there are $p$ patches, which is inconsistent. Therefore, MLP is used to align the $p_c$ to $p$. The motion vector of each patch can be calculated. $\mu, \sigma$ are utilized for normalization. $\beta, \gamma$ are used for shift and scaling during linear projection of temporal latent $\hat{z}^{(k)}$ and pose latent $z^{(k)}_p$ for $k$-th layer, respectively. The equations are as follows.

\begin{equation}
    \hat{z}^{(k)} = \frac{z^{(k)} - \mu_k}{\sqrt{\sigma_k^2 + \epsilon}}
    \label{eq:guass}
\end{equation}

\begin{equation}
    \hat{z}^{(k)}_{all} = \hat{z}^{(k)} \oplus z^{(k)}_p
    \label{eq:add}
\end{equation}

\begin{equation}
    z^{(k+1)} = \gamma_k \otimes \hat{z}^{(k)}_{all} + \beta_k
    \label{eq:norm}
\end{equation}

\subsection{Training Details and Data Processing}

\textbf{Training Details.} The  Open-Sora's second training stage is utilized to train the VAE of camera pose sequences. Specifically, the training strategy supervises the reconstruction process, including reconstruction loss and KL loss. The reconstruction loss aims to minimize the gap between the predicted result and the ground truth, while the KL loss minimizes the divergence between the VAE’s output distribution and the standard normal distribution. During the fine-tuning of the latent motion using MLP, we freeze the ST-DiT parts except for the temporal attention layer and introduce LoRA during the update of the self-attention to reduce VRAM usage. Additionally, a novelty loss function is introduced for fine-tuning, which incorporates $p_m$ as camera pose motion conditional inputs, comparing to ~\eqref{eq:loss}.
\begin{equation}
   \mathcal{L}(\theta) = \mathbb{E}_{z_0,c,t,\epsilon,p_m  }[\left\| \epsilon_\theta (z_t,c,t,p_m) - \epsilon \right\|_2 ^2]
  \label{eq:loss2}
\end{equation}

\textbf{Data Processing.} Various forms of condition input, including camera pose representation, text prompt and reference image, are carefully considered before fine-tuning. For a better camera pose representation, we randomly select 17-frame video segments and get their 12-point camera pose from timestamp information. Then we use sparse motion sampling method mentioned in Section~\ref{sec:CPA} to get the RGB image of the motion field as the camera pose representation, which gets the alignment with the sampling frame motion. For text prompt and reference image, we follow the pretrained model in OpenSora, with T5 model and 3D-VAE model, respectively.

\begin{figure*}
  \centering
  
   \includegraphics[width=0.75\linewidth]{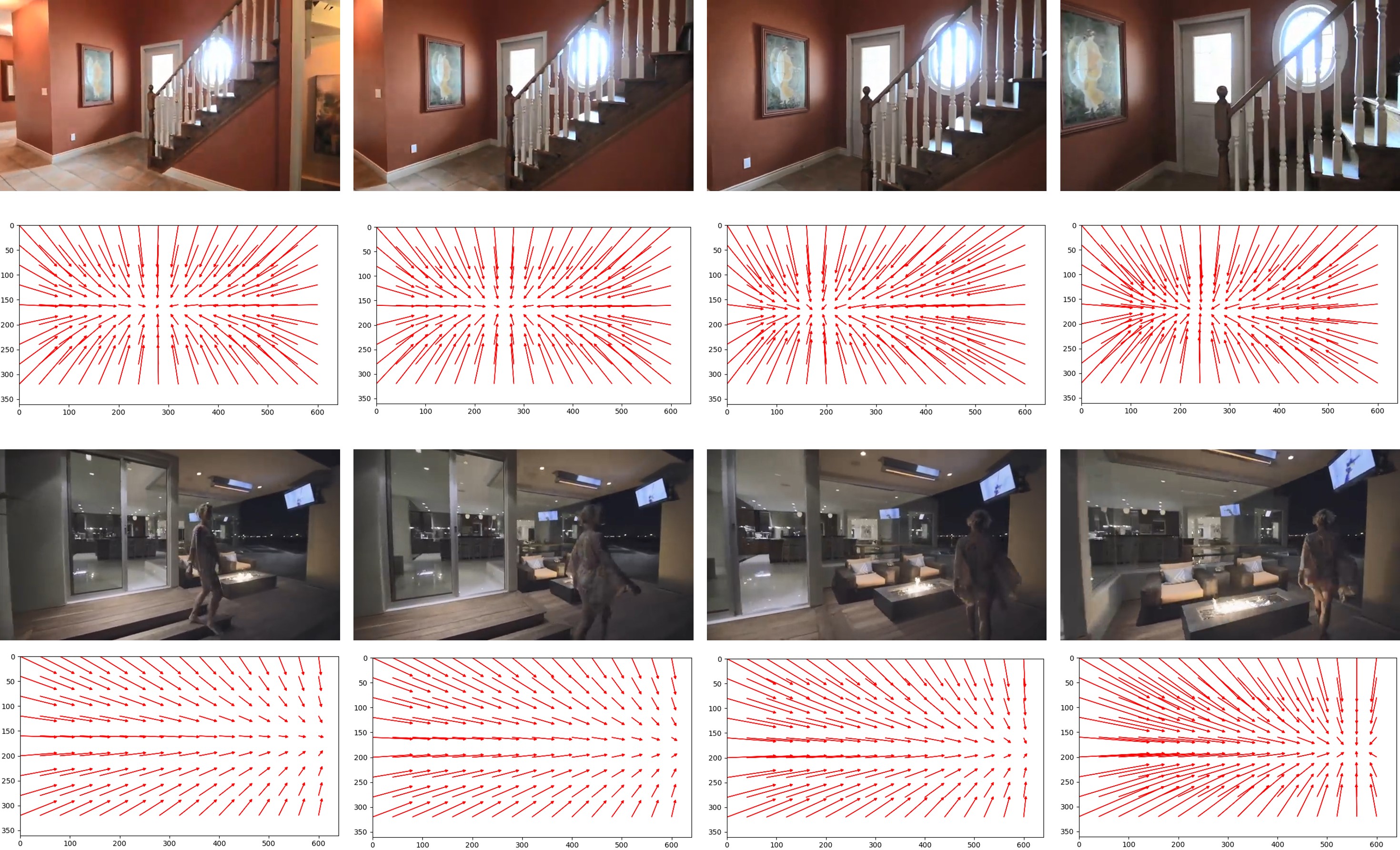}

   \caption{A visualization for camera pose series. We visualize image sequence after sparse motion sampling, with each row representing frame 0, frame 5, frame 10, and frame 15 (final frame) of the camera pose series from left to right. The arrows in the image indicate the motion of the sampling points. The first row shows a camera zoom-in motion, and the second row shows a pan-right motion.}
   \label{fig:pose_vis}
\end{figure*}

\section{Experiments}

\subsection{Implementation Details}
We initialize the weights with OpenSora-v1.2. When training the Sparse Motion Encoding Module, only the parameters of the motion-relative part and the temporal-attention part are adjusted, while the backbone is frozen to retain the original capabilities. Following the same manner as MotionCtrl ~\cite{wang2024motionctrl}, we extract 16-frame camera pose information, convert it into a RGB sparse representation (as shown in Fig.~\ref{fig:pose_vis}), and feeding it into the VAE for better reconstruction. The guidance scale is set to 7.0. The CPA is fine-tuned for 100k steps on 4 Nvidia L40s with the learning rate of  $5\times 10^{-5}$ and guidance scale of 7.0, which takes approximately 2.5 days.

\subsection{Datasets}
To validate the effectiveness of the proposed method, we use the RealEstate10K dataset, consistent with MotionCtrl and VD3D. We randomly select 20 videos from the test set, which include common camera movements such as pan left/right, up and down, zoom in/out, as well as roundabout and other complex movements.

\subsection{Metrics}

We use Fréchet Inception Distance (FID)~\cite{heusel2017gans}, Fréchet Video Distance (FVD)~\cite{unterthiner2018towards}, and CLIP Similarity (CLIPSIM)~\cite{radford2021learning} as metrics to evaluate the image quality, video consistency, and semantic similarity of the generated videos. For the camera pose consistency metric, we adopt the CamMC, the same approach mentioned in MotionCtrl. Since DiT demonstrates advantages in long video generation, we test the performance of video generation extended to 72 frames. For LDM methods, we produce long videos by using the final frame of the previous segment as the reference for the subsequent segment.

\subsection{Quantitative and Qualitative Results}

We evaluate the performance of several video generation models on both short video (16 frames) and long video (72 frames) generation tasks. The methods include LDM-based approaches such as SVD~\cite{blattmann2023stable}, AnimateDiff~\cite{guo2023animatediff}, MotionCtrl~\cite{wang2024motionctrl}, and CameraCtrl~\cite{he2024cameractrl}, and DiT-based methods like EasyAnimate~\cite{xu2024easyanimate}, VD3D~\cite{bahmani2024vd3d}, and OpenSora~\cite{opensora}, as shown in Table ~\ref{table.1}. The resolution for LDM-based methods is mainly $256 \times 256$ or $384 \times 256$, while DiT-based methods use a unified resolution of $640 \times 360$. For short video generation tasks, MotionCtrl shows an advantage, achieving the best results in video consistency metrics (FVD and CamMC). However, in long video generation tasks, CPA demonstrates significant advantages in consistency metrics. This is mainly attributed to CPA’s more precise camera pose sequences input during long video generation, which allows for fine-grained control over each frame. Additionally, it outperforms previously proposed methods in the CLIPSIM metric as well, which demonstrates that CPA effectively retains reference image. This is because we freeze other irrelevant parameters as much as possible when introducing camera pose sequences, preserving the model's original video generation capabilities.

\begin{figure*}
    \centering
    \includegraphics[width=0.8\linewidth]{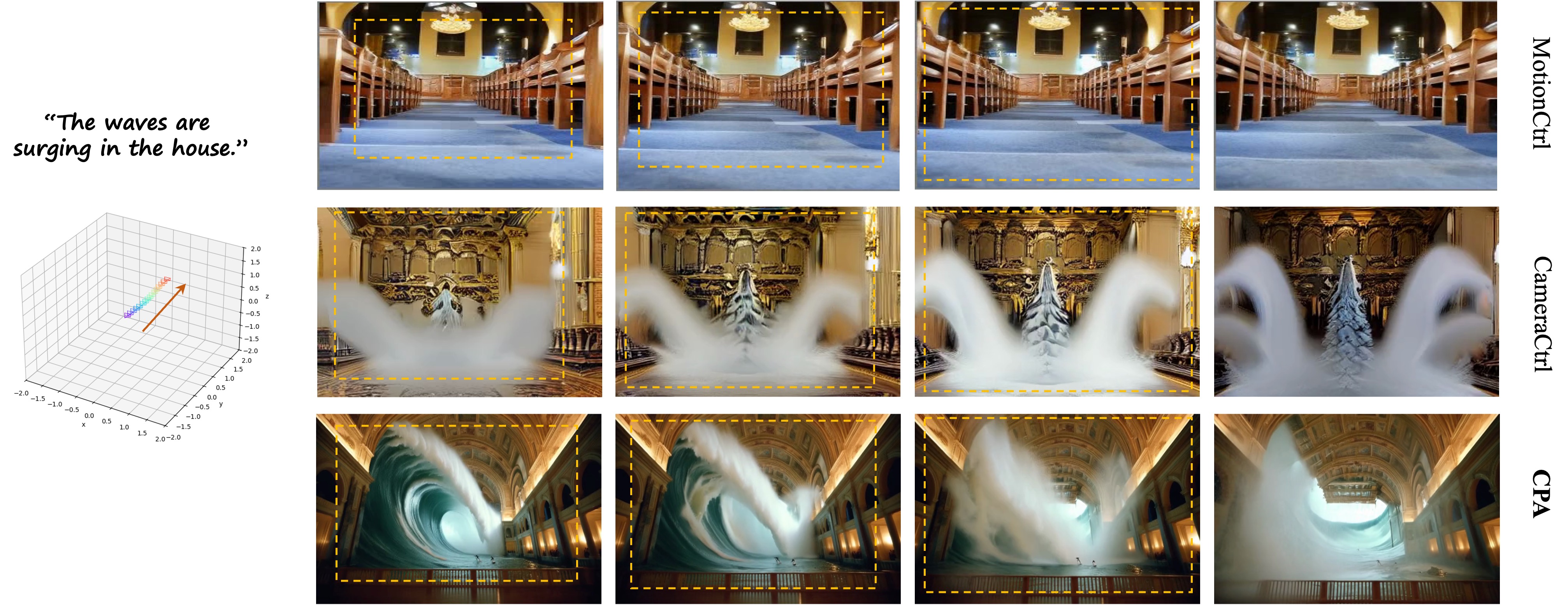}
    \caption{The performance of “zoom in” on three video generation methods, MotionCtrl, CameraCtrl and CPA. The text prompt is: “The waves are surging in the house.” The trajectory of the camera pose is shown in the 3D coordinate system, starting from the purple point to the red point. For “zoom in”, the camera position moves along the positive direction of the $y$ axis. Each row shows selected frames from the generated video. The yellow box represents the last frame range in the previous frames. All three methods demonstrate reasonable consistency in preserving camera motion. However, the text understanding of MotionCtrl and CameraCtrl is relatively poor, such as the lack of understanding of “waves”.}
    \label{fig:zoomin}
\end{figure*}

\begin{figure*}
    \centering
    \includegraphics[width=0.8\linewidth]{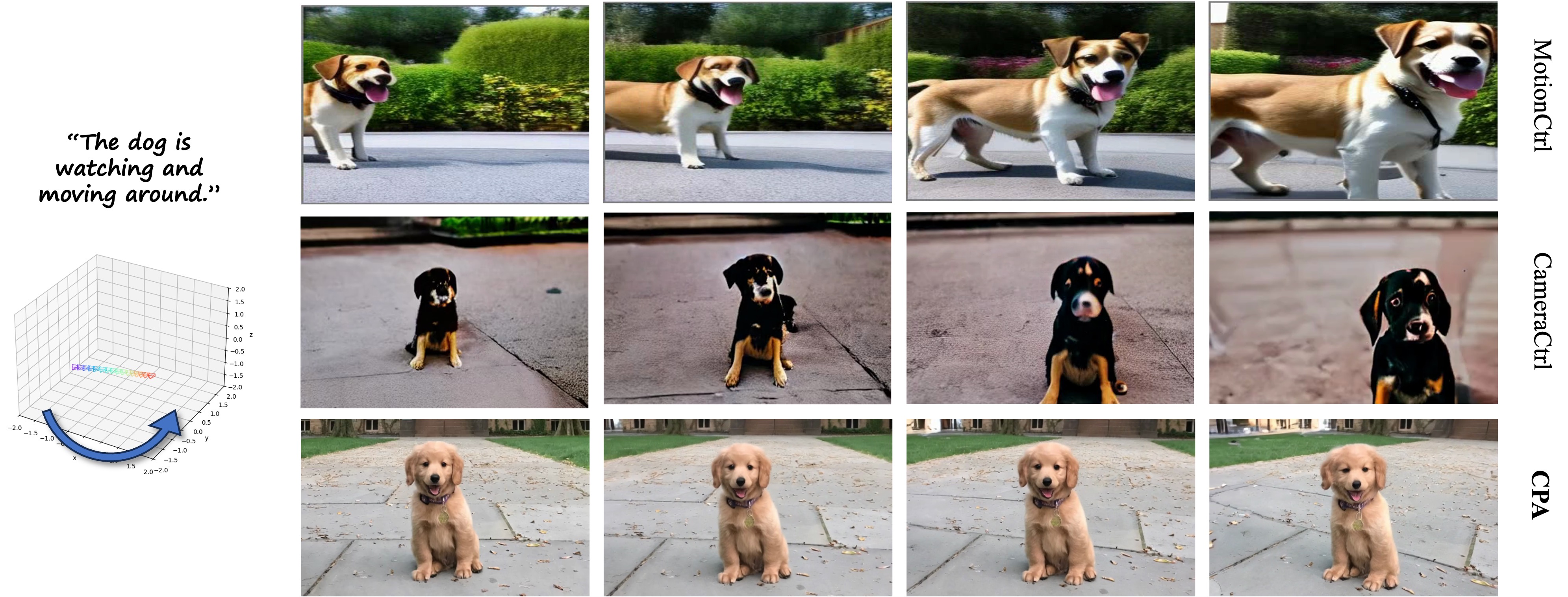}
    \caption{The performance of “roundabout” on three video generation methods, MotionCtrl, CameraCtrl and CPA. The text prompt is: “The dog is watching and moving around.” For “roundabout”, the camera's direction changes as the position moves, so the blue curve is used to represent it in the 3D coordinate system. Each row shows selected frames from the generated video. Both MotionCtrl and CameraCtrl exhibit noticeable drift of the object and struggle to achieve effective trajectory control, while CPA demonstrates more stable camera motion and object consistency.}
    \label{fig:roundabout}
\end{figure*}

\begin{figure*}
    \centering
    \includegraphics[width=0.8\linewidth]{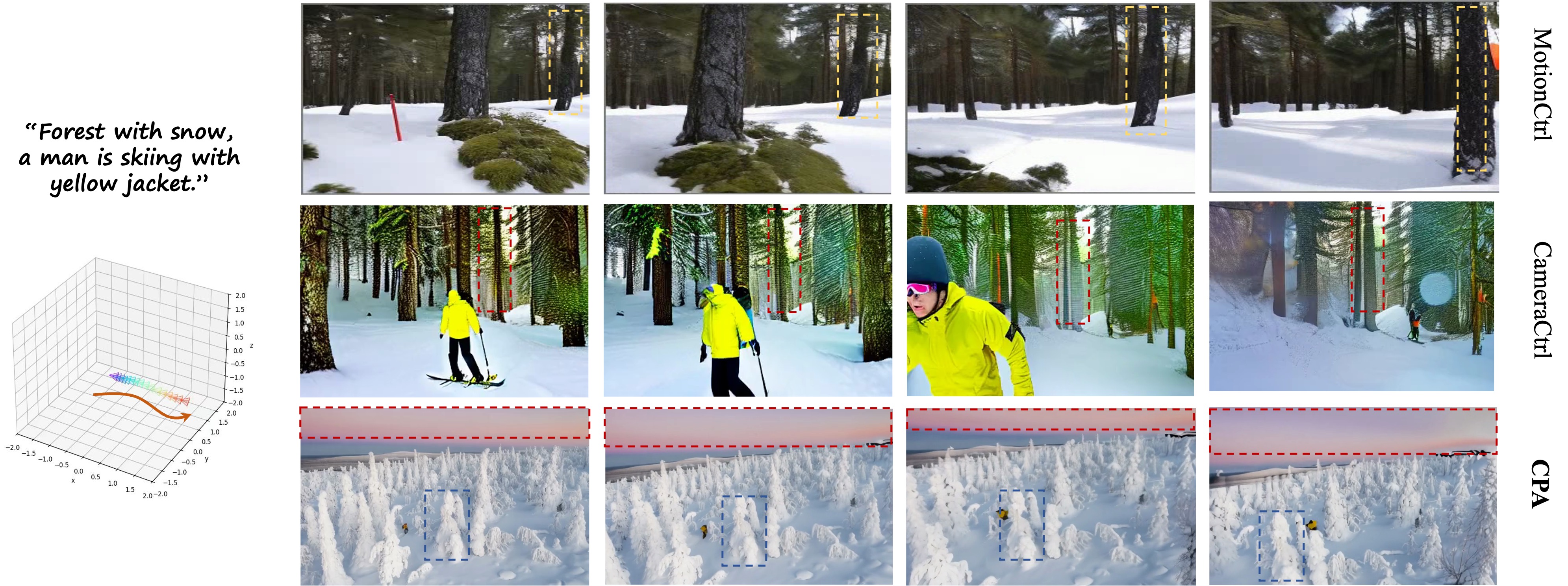}
    \caption{The performance of “shaking” camera pose on three video generation methods, MotionCtrl, CameraCtrl and CPA. The text prompt is: “Forest with snow, a man is skiing with yellow jacket.” Each row shows selected frames from the generated video. Obvious objects between frames have been marked. Both MotionCtrl and CameraCtrl have difficulty tracking under complex motion like “shaking”. MotionCtrl fails to understand the text prompt “man”. CameraCtrl has only part of the frames on the “man”. CPA retains the shaking camera trajectory well~(red box) while retaining the detailed information of the “man”, showing good object motion effects.}
    \label{fig:shake}
\end{figure*}

We also present the visualized performance of video generation using CPA (Fig.~\ref{fig:zoomin},~\ref{fig:roundabout} and~\ref{fig:shake}). For simple camera pose, such as “zoom in” and “roundabout”, CPA performs excellently on these basic camera movement tasks, accurately following the camera motion poses. For complex tasks like “shaking”, CPA achieves smooth transitions while maintaining the object motion effectively.

\begin{table*}[]
	\renewcommand{\arraystretch}{1.2}
	\centering
	\caption{Comparison of consistency performance using different video generation methods, our method CPA achieves the best results in long video task.}
	\scalebox{0.8}{\setlength{\tabcolsep}{2.5mm}{ 
		\begin{tabular}{c|cc|cc|cc|cc}
			\toprule
			\multirow{2}{*}{\centering \textbf{Models}} & \multicolumn{2}{c|}{\textbf{FID ($\downarrow$)}} & \multicolumn{2}{c|}{\textbf{FVD ($\downarrow$)}} & \multicolumn{2}{c|}{\textbf{CLIPSIM  ($\uparrow$)}} & \multicolumn{2}{c}{\textbf{CamMC  ($\downarrow$)}}  \\ \cline{2-9}
			 & \multicolumn{1}{c}{\textbf{Short}} & \multicolumn{1}{c|}{\textbf{Long}} & \multicolumn{1}{c}{\textbf{Short}} & \multicolumn{1}{c|}{\textbf{Long}} & \multicolumn{1}{c}{\textbf{Short}} & \multicolumn{1}{c|}{\textbf{Long}} & \multicolumn{1}{c}{\textbf{Short}} & \multicolumn{1}{c}{\textbf{Long}}  \\ 
   \midrule
			SVD~\cite{blattmann2023stable}                & 185 & 261 & 1503 & 1628 & 0.1604 & 0.1102 & 0.160 & 0.885  \\
			AnimateDiff~\cite{guo2023animatediff}         & 167 & 175 & 1447 & 1512 & 0.2367 & 0.2045 & 0.051 & 0.473  \\
			
            MotionCtrl~\cite{wang2024motionctrl}              & \textbf{132} & 168 & \textbf{1004} & 1464 & 0.2355 & 0.2268 & \textbf{0.029} & 0.472  \\
			CameraCtrl~\cite{he2024cameractrl}            & 173 & 254 & 1426 & 1530 & 0.2201 & 0.2194 & 0.052 & 0.205  \\ \midrule
			EasyAnimateV3~\cite{xu2024easyanimate}        & 165 & 245       & 1401 & 1498 & 0.2305 & 0.2250 & 0.046 & 0.068  \\
			VD3D~\cite{bahmani2024vd3d}                  & -- & 171       & -- & 1400 & -- & 0.2032 & -- & 0.044  \\
                
            OpenSora~\cite{opensora}                         & 141 & 161       & 1587 & 1682 & 0.2496 & 0.2284 & -- & --  \\
			CPA (Ours)                               & 147 & \textbf{158}  & 1310 & \textbf{1387} & \textbf{0.2521} & \textbf{0.2438} & 0.037 & \textbf{0.042}  \\ \bottomrule
		\end{tabular}
    }}
	\label{table.1}
\end{table*}

\subsection{Ablation Studies}
We conduct ablation studies for CPA, focusing on the sampling interval of camera pose RGB series and the temporal injection methods, corresponding to the Sparse Motion Encoding and Temporal Attention Injection Module introduced in Section~\ref{sec:CPA}.

In the sampling interval experiment, we conduct three sets of motion extraction strategies: $20\times$, $40\times$, and $80\times$. For example, for $640\times 360$ video resolution, the $40\times$ strategy corresponds to $16\times 9$ motion extraction points. We train the VAE using different sampling strategies and evaluate the video generation performance, as shown in Table~\ref{table.2}. We find that the $40\times$  achieves the best results across all metrics, indicating that the camera pose motion sampling quantity at $40\times$ is relatively optimal. For the $20\times$ and $80\times$, we observe varying degrees of target drift or weakened motion consistency during evaluation. The possible reason is that for $80\times$, the sampling density is sparse (around 40 vectors per frame), making it easy for targets to be distorted and reducing motion control capability. On the other hand, for $20\times$, there are over 500 vectors each frame, making it difficult to align with each motion vector and leading to a decrease in motion consistency. This ablation study provides a reference for quantifying sparse motion sampling.

In the injection method experiment, we also use three strategies: channel-dimension concatation (concat), cross-attention and our injection module~(TAI). Channel-dimension concatation adds the camera pose motion latent to the temporal layers, which is used in MotionCtrl. For cross attention, temporal layers represents query, while latent motion
represents key and value, calculates the hidden layers.  The video generation performance for the three methods are shown in Table~\ref{table.3}. We find that TAI achieves better consistency results compared to the other methods.  The reason is that for channel-dimension concatation, which fails to align the motion latent with the temporal attention at first, leading to weaker camera pose control during generation. For cross-attention, which  alters the dimension of both motion latent and temporal attention, causes more disruption to the temporal attention in the original network. Additionally, we observe that our method is able to unify pose and temporal latent into a similar distribution, which is crucial for the effective injection of camera pose.

\begin{table}[tb!]
    \renewcommand{\arraystretch}{1.2}
    	\centering
    	\caption{Ablation study results showing the effect of sample ratios for camera pose latents.}
    	\scalebox{0.8}{\setlength{\tabcolsep}{2.5mm}{ 
    		\begin{tabular}{c|cccc}
                \toprule
                \textbf{Ratios} & \textbf{FID ($\downarrow$)} & \textbf{FVD ($\downarrow$)} & \textbf{CLIPSIM ($\uparrow$)} & \textbf{CamMC ($\downarrow$)} \\ 
                \midrule
                $20\times$ & 156& 1395& 0.2328&0.045\\
                $40\times$ &  \textbf{148} &  \textbf{1313} &  \textbf{0.2521} &  \textbf{0.038}\\
                $80\times$ & 151& 1358& 0.2462& 0.042 \\
                \bottomrule
            \end{tabular}
        }}
    \label{table.2}
    
\end{table}

\begin{table}[tb!]
    \renewcommand{\arraystretch}{1.2}
    	\centering
    	\caption{Ablation study results showing the effect of different injection modules for camera pose latents.}
    	\scalebox{0.8}{\setlength{\tabcolsep}{2mm}{ 
    		\begin{tabular}{c|cccc}
                \toprule
                \textbf{Methods} & \textbf{FID ($\downarrow$)} & \textbf{FVD ($\downarrow$)} & \textbf{CLIPSIM ($\uparrow$)} & \textbf{CamMC ($\downarrow$)} \\ 
                \midrule
                Concat & 152 & 1342 & 0.2328 & 0.046\\
                
                Cross Attn. &  149 & 1326 & 0.2335 & 0.041\\
                TAI &  \textbf{148} &  \textbf{1313} &  \textbf{0.2521} &  \textbf{0.038} \\
                
                \bottomrule
            \end{tabular}
        }}
    \label{table.3}
    
\end{table}

\subsection{Discussions}

\begin{figure*}
    \centering
    \includegraphics[width=0.8\linewidth]{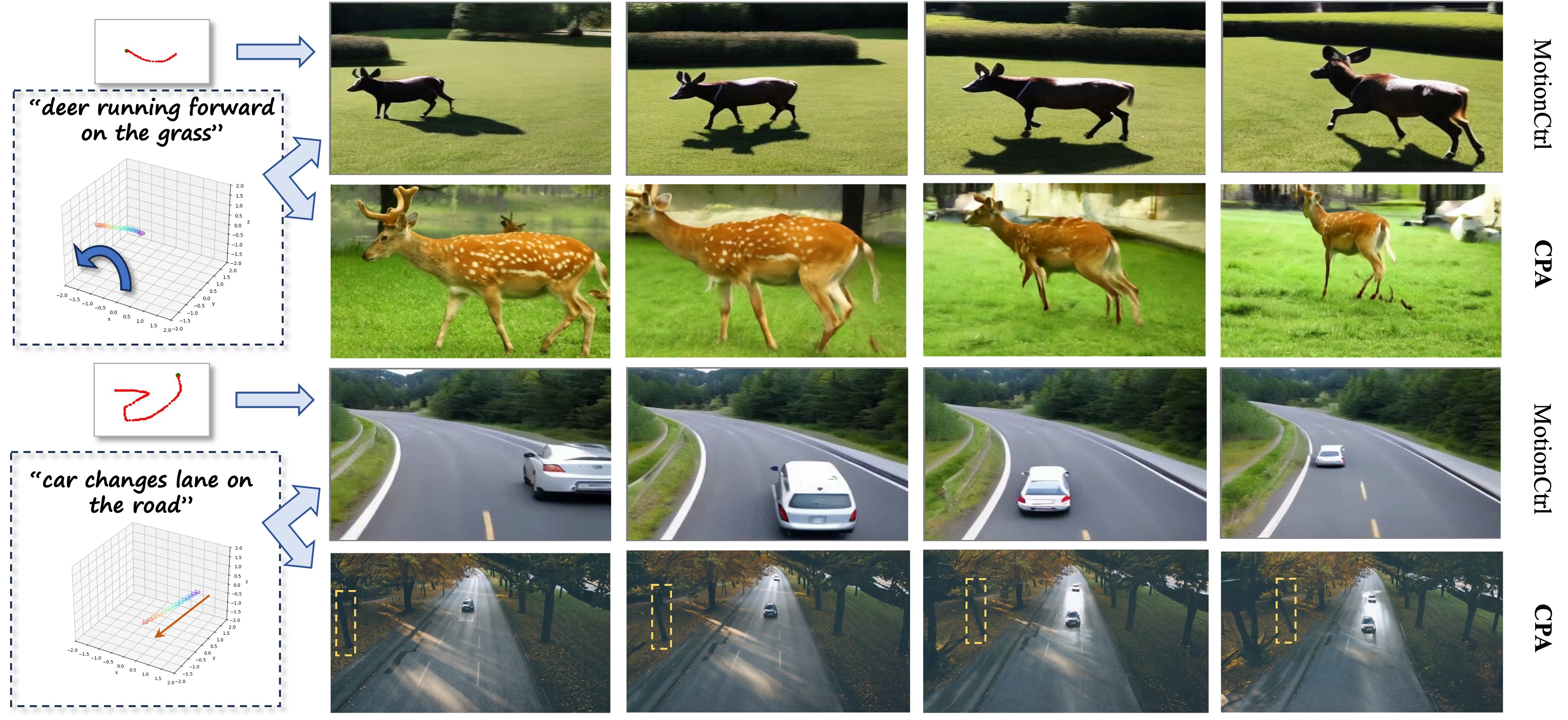}
    \caption{The performance of MotionCtrl with object motion and CPA without object motion.  Two cases are used, one with simple object motion and complex camera pose, and the other with complex object motion and simple camera pose. MotionCtrl has three inputs: object motion, camera pose, and text prompt, while CPA has only two inputs: camera pose and text prompt. The green dot is used as the starting point of the object motion. Each row shows selected frames from the generated video. The two sets of experimental results show that, as the object motion becomes more complicated, MotionCtrl cannot handle both the camera pose and the object motion well. CPA can ensure the rationality of the object motion while following the camera pose. However, CPA tends to be more conservative in object motion.}
    \label{fig:obj}
\end{figure*}

CPA demonstrates excellent performance in maintaining camera pose consistency for long video generation, but there are still the following challenges and limitations:

\begin{itemize}
    \item \textbf{The performance of the object consistency is relatively weak.} The consistency of camera pose motion is mainly considered in CPA. Regarding to object consistency, we conduct a comparative experiment between CPA and MotionCtrl. The results are shown in the Fig.~\ref{fig:obj}. Although object consistency is also preserved, due to the conservative nature of motion estimation, the object movement tends to be limited to small-scale motions, making large-scale motion generation more challenging.
    \item \textbf{There is limited support for camera pose motion trajectories.} To ensure consistency in our study, we use camera pose condition based on 16 frames. More frame requirements rely on frame interpolation for completion. Currently, generating more complex motion videos remains a challenge.
\end{itemize}

\section{Conclusion}
We propose a novelty method for camera-pose-awareness video generation based on DiT architecture. To effectively inject camera pose sequences into the temporal-attention layer, we introduce a Sparse Motion Encoding Module and Temporal Attention Injection Module that transforms motion into sampling points in the RGB space and use layer normalization and MLP to achieve pose latent embedding. Our method achieves SOTA in camera motion control for long video generation.

{\small
\bibliographystyle{ieee_fullname}
\bibliography{egbib}
}

\end{document}